\documentclass[letterpaper]{article} % DO NOT CHANGE THIS
\usepackage{aaai25}  % DO NOT CHANGE THIS
\usepackage{times}  % DO NOT CHANGE THIS
\usepackage{helvet}  % DO NOT CHANGE THIS
\usepackage{courier}  % DO NOT CHANGE THIS
\usepackage[hyphens]{url}  % DO NOT CHANGE THIS
\usepackage{graphicx} % DO NOT CHANGE THIS
\usepackage{natbib}  % DO NOT CHANGE THIS AND DO NOT ADD ANY OPTIONS TO IT
\usepackage{caption} % DO NOT CHANGE THIS AND DO NOT ADD ANY OPTIONS TO IT
\usepackage{algorithm}
\usepackage{algorithmic}
\usepackage{amsmath}
\usepackage{newfloat}
\usepackage{listings}
\usepackage{url}
\usepackage{multirow}
\usepackage{graphicx}
\usepackage{subcaption}
\usepackage{amssymb}

\DeclareCaptionStyle{ruled}{labelfont=normalfont,labelsep=colon,strut=off}

\floatstyle{ruled}
\newfloat{listing}{tb}{lst}{}
\floatname{listing}{Listing}

\title{Irregularity-Informed Time Series Analysis: Adaptive Modelling of Spatial and Temporal Dynamics}

\author{
Liangwei Nathan Zheng\textsuperscript{1}, 
Zhengyang Li\textsuperscript{1}, 
Chang George Dong\textsuperscript{1},
Wei Emma Zhang\textsuperscript{1}, 
Lin Yue\textsuperscript{1}, 
Miao Xu\textsuperscript{2}, 
Olaf Maennel\textsuperscript{1}, 
Weitong Chen\textsuperscript{1}
}

\affiliations {
The University of Adelaide\textsuperscript{1}, The University of Queensland\textsuperscript{2} \\
\{liangwei.zheng, chang.dong, wei.e.zhang, lin.yue, t.chen\}@adelaide.edu.au, xin.chen@nottingham.ac.uk
}

\usepackage{bibentry}
\begin{document}
\maketitle

\begin{abstract}
  Irregular Time Series Data (IRTS) has shown increasing prevalence in real-world applications. We observed that IRTS can be divided into two specialized types: Natural Irregular Time Series (NIRTS) and Accidental Irregular Time Series (AIRTS). Various existing methods either ignore the impacts of irregular patterns or statically learn the irregular dynamics of NIRTS and AIRTS data and suffer from limited data availability due to the sparsity of IRTS. We proposed a novel transformer-based framework for general irregular time series data that treats IRTS from four views: Locality, Time, Spatio and Irregularity to motivate the data usage to the highest potential. Moreover, we design a sophisticated irregularity-gate mechanism to adaptively select task-relevant information from irregularity, which improves the generalization ability to various IRTS data. We implement extensive experiments to demonstrate the resistance of our work to three highly missing ratio datasets (88.4\%, 94.9\%, 60\% missing value) and investigate the significance of the irregularity information for both NIRTS and AIRTS by additional ablation study. We release our implementation in \url{https://github.com/IcurasLW/MTSFormer-Irregular_Time_Series.git}.
\end{abstract}

\section{Introduction}
\begin{figure}
    \centering
    \includegraphics[width=0.9\linewidth]{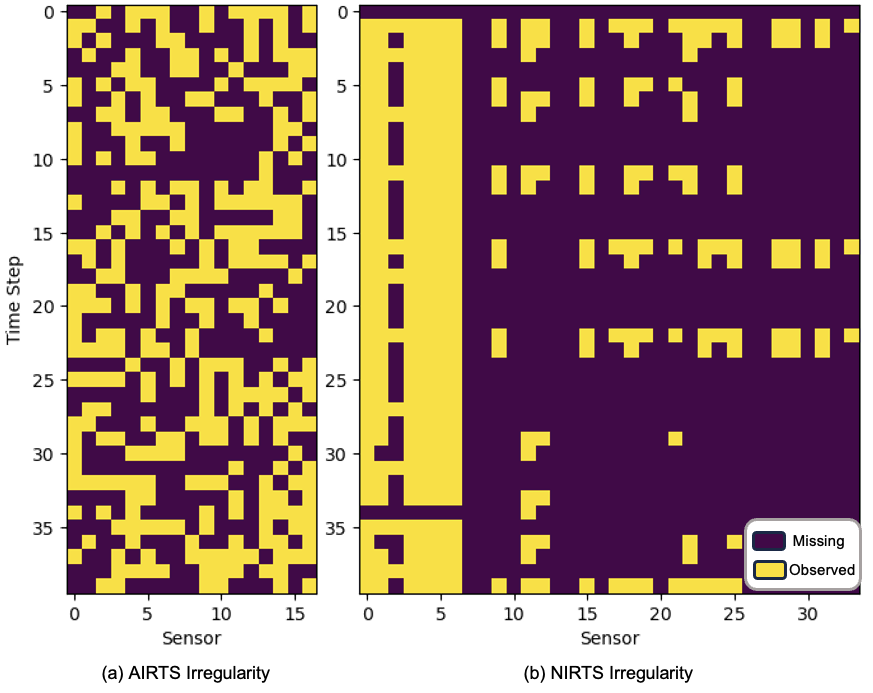}
    \caption{(a) AIRTS irregularity pattern from PAM dataset (b)NIRTS irregularity pattern from P19 dataset}
    \label{fig:NIRTS and AIRTS}
\end{figure}

Multivariate Time Series data is commonly observed in real-world applications such as medical records, traffic, and engineering industry \cite{weerakody2021review,chen2024contiformer,shen2022death,xu2024reliable,yue2021intention,zhao2021telecomnet,qiu2023pre,dong2023swap}. In practice, missing values, local sensor failure of the system and inconsistent sampling frequencies often lead to Irregular Time Series (IRTS) data, forming by non-uniform time intervals between successive measurements \cite{schirmer2022modeling}. Many real-world irregular datasets, especially in the medical domain, are extremely sparse and the irregularity is informative for prediction \cite{weerakody2021review}. There are two categories of data irregularity: Natural Irregular Time Series (NIRTS) due to different sampling rates of measurements, and Accidental Irregular Time Series (NIRTS) due to local failure. These two IRTS data types often lead to sparse and non-uniform measurements, resulting in large missing value proportions in time series as shown in Figure. \ref{fig:NIRTS and AIRTS}. NIRTS often displays indicative patterns, which can reveal underlying missing patterns \cite{sun2020review,kreindler2016effects}. For example, medical sensors in hospitals record data at varying frequencies and intervals. A patient with life-threatening conditions may have data recorded more frequently, particularly on indicators related to their health status. AIRTS, on the other hand, usually displays random missing patterns or disconnections during a period. For example, a traffic sensor may shut down due to an electrical outage or poor contact, medical sensor may be inactive due to mechanical faults or the absence of devices. Such accidental missing will not contribute to prediction and often degrade the performance of models.

Typical sequential models (e.g. LSTM, GRU) generally assume the data are regular \cite{zhang2021graph,siami2018comparison} and fill the missing values between two effective measurements with interpolation and statistical approaches to construct regular time series \cite{chen2023death,rubanova2019latent}. Previous studies \cite{rubanova2019latent,oskarsson2023temporal} also attempted to model the dynamic continuity of irregular data by Ordinary Differential Equation (ODE) Network, which assumes the discrete and sparse data observation can be represented by continuous latent dynamics. However, this temporal approach does not exploit the rich information in missing patterns and generalizes poorly in NIRTS since they ignore the informative missing patterns. Moreover, the training process of the ODE family models is computationally expensive on time complexity since the ODE solver sequentially solves the time dynamics by CPU \cite{bilovs2021neural,norcliffe2023faster}.

Some studies \cite{zhang2021graph,oskarsson2023temporal,wu2021dynamic,zhang2019deep,little2019statistical} noted that the inter-sensor correlations between observed and unobserved sensors can be utilized to model the pattern of inactivate sensors by message passing mechanism of Graph Neural Network (GNN). They respect the missing nature of IRTS and learned the sensor dependency and the correlation between sensors by message passing mechanism \cite{zhang2021graph, wu2020connecting}. However, the message-passing mechanism for sparsely observed sensors essentially imputes the inactive sensor by the active sensor using linear or non-linear transformation. This approach does not generalize well to sparsely observed datasets since the fewer sampling sensors will be dominated by the information of frequent sampling sensors, resulting in similar representation sensors and sacrificing the unique characteristics of those less frequented sampled sensors. The graph approaches also assume that the input graph is fully connected or requires prior knowledge to construct graph dependency, limiting the model from general applications in the real world.

Recently, Transformer-based models \cite{nie2022time,liu2023itransformer} demonstrated efficacy in modelling time series data for forecasting and classification tasks. Nevertheless, these methods assume the time series data is regular and fully observed, thereby rendering the forecasting techniques challenging to adapt to our highly sparse IRTS data. Although there are some works \cite{zhang2023warpformer, zhang2023warpformer,tipirneni2022self} utilizing the Transformer backbone for IRTS, they also ignored the information from irregular patterns such as sampling frequency and sampling density, which may introduce measurement bias from the original value observation due to a wider range of measurement value. For example, the normal respiratory rate and heart rate vary largely in different ages, genders, and pregnant groups \cite{shen2022leads}. Instead of focusing on the observed measurement, we hypothesize that the irregularity patterns are sufficiently informative on modelling NIRTS and negatively contribute to the AIRTS. The primary reason is that AIRTS has no indicative but noisy missing patterns that distract the model from learning the actual distinct feature. We show an illustration of AIRTS and NIRTS from a real-world IRTS dataset in Figure. \ref{fig:NIRTS and AIRTS}, NIRTS usually exhibits a clear sampling pattern of each sensor while the irregularity of AIRTS is noisy and distractive.

In addition, Many existing works on learning from IRTS focus solely on the time dimension, neglecting interactions from other dimensions such as spatial and locality information. This narrow approach limits the models' effectiveness when dealing with the sparse and irregularly observed data in IRTS. Moreover, these works often overlook irregularity or simply concatenate an irregularity mask to the original data, failing to account for the general characteristics of IRTS datasets. As a result, they may provide sub-optimal solutions across different types of IRTS data. None of the previous studies offers a unifying approach that generalizes well to both types of IRTS datasets. To this end, we identify two key challenges in modelling IRTS:

\textbf{\underline{Suffering from Limited Observed Data} : }
Many real-world data such as the medical domain often display a high missing rate (e.g. 94.9\% missing data for P19 dataset) and limited measurement in a period of time. Previous studies only consider the information from the time steps view and obtain sub-optimal solutions due to the limited observed data.

\textbf{\underline{Reliability to General IRTS} : }
Simple concatenation or neglecting the irregular nature can lead to poorly generalized models for both NIRTS and AIRTS datasets, thereby constraining the ability to handle diverse IRTS data. Existing methodologies excel on one dataset while performing inadequately on the other.

\begin{figure*}[ht]
    \centering
    \includegraphics[scale=0.28]{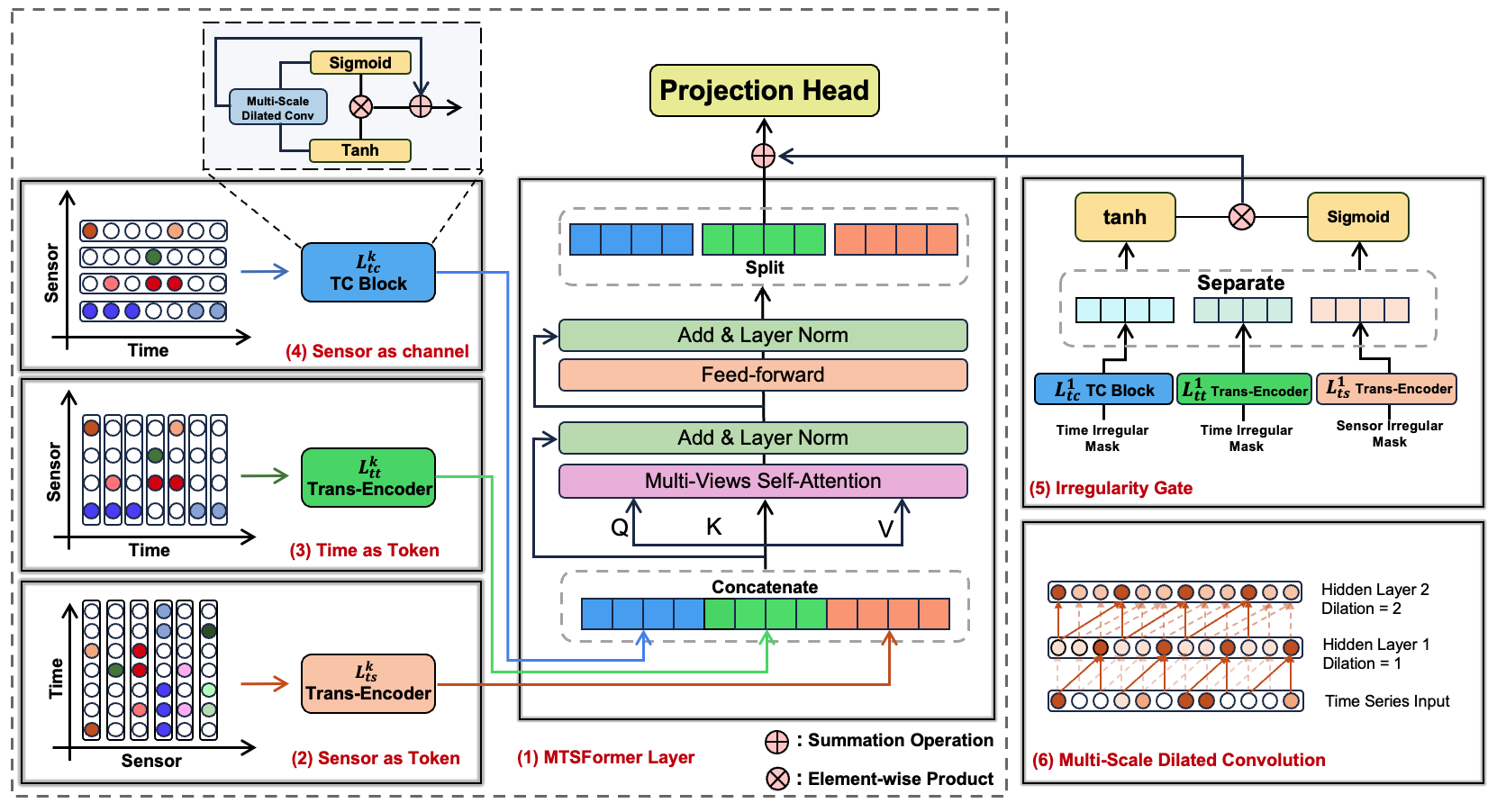}
    \caption{MTSFormer Overall Architecture: (1) MTSFormer with $\mathcal{K}$ layer MTSFormer block accepts three different views as inputs: Local, Time and Sensor. Local, Time and Sensor views are encoded by three individual encoders $L^k_{tc}$, $L^k_{tt}$ and $L^k_{ts}$.  (2) Sensor as token path transposes the feature dimension of $x^t_i$ to the reading of time dimension, focusing on the interaction between sensors. The encoder of of Sensor as token path is a one-layer vanilla transformer encoder. (3) Time as Token considers the time steps as a token where the sensor readings in certain time steps are the features. The encoder of Time Token $x^t_i$ is a one-layer vanilla Transformer encoder. (4) Sensor as Channel inputs the same data $x^t_i$ as Time as Token path. TC block represents a Temporal Dilated Convolution block with multi-scale dilated rate that extracts the locality information of time series, focusing on detailed and temporal information. (5) Irregularity gate is encoded by the first layer encoders of each path from MTSFormer in a share-weight manner, $L^1_{tc}$, $L^1_{tt}$ and $L^1_{ts}$ and pass the irregularity embedding into a gate activation function. Tanh maps the information to $[-1, 1]$ and sigmoid maps the embedding to $[0, 1]$, which allows the model adaptively preserve useful information from irregularity. (6) Multi-scale dilated convolution with different dilation rates captures the local irregular patterns by the large receptive field.}
    \label{fig:overall-architecture}
\end{figure*}

To address the above challenges, we propose Multi-View Transformer for General Irregular Time Series Data (MTSFormer) as shown in Figure \ref{fig:overall-architecture}, motivating IRTS data utilization to its fullest potential by leveraging information from different views. Our intuition is that learning the multi-variate irregular time series can fuse information from local, global and irregularity views. Local views focus on details but distinct features such as the local irregular patterns around an effective measurement. Global views focus on low-frequency features such as tendency and periodicity. Irregular patterns contain information such as sampling frequency and interval, which is significantly indicative for NIRTS, but not less informative and distractive to AIRTS. To generalize on both NIRTS and AIRTS data, we design an irregular gated mechanism to adaptively control the information flow from irregular patterns without prior knowledge of the dataset. In addition, we implement extensive experiments to demonstrate the effectiveness of our proposed method and SOTA methods in different missing ratios of sensors. We extensively discuss the impacts of irregularity patterns and different components on NIRTS and AIRTS by ablation study. 

The contributions of our work in leveraging Transformer-based models for irregular time series data can be summarized as follows:

\begin{itemize}
    \item \textbf{Enhancing Model Performance with Irregularity Patterns:} Our research leverages irregularity patterns to provide a unifying method to model both NIRTS and AIRTS data. Additionally, it broadens data integration by incorporating spatial and locality dimensions, filling gaps left by previous models.

    \item \textbf{Innovative Multi-View Transformer Design:} We proposed a novel Multi-View Transformer that processes data from both local and global perspectives. This also is enhanced by an adaptive irregular gated mechanism that optimizes the use of irregular patterns to maximize informative content and minimize noise.

    \item \textbf{Extensive Validation and Impact Analysis:} Our experiments and ablation studies demonstrate the effectiveness of our approach across diverse datasets, validating its superiority in handling irregular time series data compared to existing methods.
\end{itemize}

\section{Related Works}

\textbf{Learning from Temporal IRTS. }
RNNs family models (e.g. GRU and LSTM) are powerful sequential models for time series data, which usually learn the model based on previous input \cite{cao2018brits,harutyunyan2019multitask}. To model irregular time series data, many RNN-based models usually preprocessed the irregular data into regular data by imputation and interpolation \cite{harutyunyan2019multitask,mozer2017discrete}. IP-Net\cite{shukla2019interpolation} incorporated sampling intensity information and GRU cells to impute IRTS back to irregular time series. Imputation may distort the underlying irregularity patterns and lead to unexpected values to degrade performance. To avoid the preprocessing, recent RNN-based models such as GRU-D \cite{che2018recurrent} and ODE-RNN \cite{rubanova2019latent} considered irregularity dynamics using the decay mechanism and continuous ordinary differential equation. Although GRU-D and ODE-RNN have been shown to outperform over vanilla RNN, the sequential modelling fails to grasp the global view of irregularity. Attention mechanisms have demonstrated the capability to aggregate global information from other tokens.  \cite{vaswani2017attention}. The computing efficiency of RNNs family model also limited the generalization ability to large-scale datasets.

\textbf{Attention-based Designs. }
Recent studies have explored various approaches for learning irregularity in data. mTAND \cite{shukla2021multi} developed a time attention mechanism to learn temporal similarity using time embedding. STraTS \cite{tipirneni2022self} combined time, value and sparse mask to form contextual embedding and fused the information by multi-head attention. WarpFormer \cite{zhang2023warpformer}, similarly, leveraged Dynamic Time Warping (DTW) into transformer architecture that unifies non-uniform time series, producing multi-scale embeddings. ContiFormer \cite{chen2024contiformer} integrated continuous ODE into the attention mechanism to calculate the latent similarity between non-uniform observations. However, these methods often overlook irregularity patterns and fail to consider indicative irregularities, which can impact model performance.

\textbf{Internal Dependency Modelling. }
There has been a surge of interest in applying Graph Neural Networks to discover the inter-sensor relationship between the non-uniform measurements. MTGNN \cite{wu2020connecting} leverage message passing network and inception temporal convolution to generate spatial-temporal representation of sensor while Raindrop \cite{zhang2021graph} models the graph dependency between irregular sensor by attention and message passing mechanism. However, learning graph dependency Raindrop usually degrades the performance of the model as it may fail to model the actual dependency of sensors in the real world. Best performance of Raindrop is obtained without graph dependency learning. Oskarsson J et al. \cite{oskarsson2023temporal} incorporates message passing with neural ODE for continuous modelling.

\textbf{Diverse Ideas for IRTS. }
In addition, SeFT \cite{horn2020set} transferred IRTS into a set of triplet representations by set function and encoded the representation using the attention mechanism. ViTST \cite{li2024time} converted time series data into time series images and utilized the powerful reasoning ability of pre-trained large vision models (e.g. ViT, Swin) to classify IRTS, resulting impressive performance over baseline. Nevertheless, the experiments of ViTST indicated that the performance originated from the pertaining of large datasets and large-scale parameters of ViT models. With the same model parameters trained from scratch, the performance is worse even than a one-layer vanilla Transformer. 

Although there are existing works exploring the model with different approaches (e.g. RNNs-based, attention-based, graph-based, large model). There are still fewer discussions on utilizing the interaction between local and global and leveraging the power of IRTS irregularity to deal with irregular time series data. To address the challenges of sparse information on IRTS and learn from irregularity, MTSFormer utilizes local, global and irregular information to generate different information paths. Our proposed method requires no prior knowledge of the dataset and can be applied to any general IRTS dataset.

\section{Proposed Method}
\subsection{Problem Definition}
The problem of interest in this work is to learn a supervised model from sparse and non-uniform observed time series data. IRTS data can be divided into two categories: Naturally Irregular Time Series (NIRTS) and Accidentally Irregular Time Series (AIRTS). NIRTS is commonly observed in medical electronic records, containing underlying and rich information in irregularity patterns. AIRTS may be caused by unexpected faults of measurements containing random irregularity and misleading information to downstream tasks. We give a notation definition as follows.

Let $ \mathcal{D}=\{  (\mathcal{X}_i, y_i) | i=1,\dots N \}$ denote a multi-variate irregular time series dataset with $N$ classes, containing sets of sample $\mathcal{X}_i$ assigned to supervision signal $y_i$. Each sample $x_i \in \mathcal{X}_i$ is a multivariate irregular time series with $\mathcal{S}$ non-uniformly observed sensors across the observed time length $T$. An unknown type of IRTS dataset is given, where each sample $x_i$ has non-uniform measurements over time and sensor dimension. The task is to learn a function $\mathcal{M}:x_i \xrightarrow{\mathcal{M}} z_i$, mapping the IRTS sample $x_i$ into a fixed-length representation $z_i$ for problems of interest, such as classification, imputation and regression. MTSFormer aims to learn the distinct irregularity and internal relationship among different views of IRTS to generate a comprehensive fused embedding $z_i$ for an input sample $x_i$.

\subsection{Power of Trinity}
IRTS is highly sparse with non-uniform measurement over time. Many existing works suffer from the limited observed data since they only consider time series from either temporal or spatial view. Considering only one perspective fails to fully capture the intricate relationship between time and variables. To motivate the data usage to its fullest potential, we consider four views of IRTS: Sensor as Channel, Time as Token, Sensor as Token and Irregularity as shown in Figure. \ref{fig:overall-architecture}. We adopt the Transformer backbone architecture for our proposed method. The key idea of MTSFormer is to fuse the information from different views of IRTS and adaptively incorporate the irregularity patterns. Specifically, the encoder of Sensor as Channel, Time as Token and Sensor as Token are individual to learn the unique characteristics of each view and map them into the same dimension space. The irregularity encoders are shared with the first layer encoder of three paths to eliminate the biases caused by inconsistent measurements.

\textbf{Time as Token:}
MTSFormer accepts a series of data $x_i$ and labels $y_i$ as input. We denote $x^t_{i, s}$ as a data observation at time step $t$ of sensor $s$ and integrate $x^t_{i, s}$ along sensor dimension to obtain a time token $x^t_{i} \in \mathbb{R}^{L \times N_s}$, where $L$ and $N_s$ are the sequential length of time step and the number of sensors respectively. $x^t_i$ will be sent to a one-layer vanilla transformer $L^k_{tt}$ encoder to embed global information over the whole time series. The output from $L^k_{tt}$ is denoted as time dimension embedding $e^t_{i} \in \mathbb{R}^{L \times E} $, where $E$ is the aligned embedding dimension for all views. The attention mechanism has a global receptive field over the sequence dimension \cite{vaswani2017attention,chen2019damtrnn}, enabling this path to capture high-level information from the time dimension such as periodicity and tendency as consistent as previous work. 

\textbf{Sensor as Token:} Similarly, we integrate $x^t_{i, s}$ along time dimension to obtain $x^s_{i} \in \mathbb{R}^{N_s \times L}$ denoted as the sensor token. $x^s_i$ is fed into another individual one-layer vanilla transformer encoder. This path also utilizes the power of self-attention to find the spatial correlations among sensors and their dependencies, mapping $x^s_i$ into informative sensor embedding $e^s_{i} \in \mathbb{R}^{N_s \times E}$. Each sensor is characterized as the property of time series with rich information in spatial dimensions. This sensor view provides a wider receptive field over the timeline as an extreme case of convolution, providing a global view of sensor interaction. The self-attention mechanism and the feedforward layer together maximize the mutual information among sensors, facilitating spatial learning. 

\textbf{Sensor as Channel:} Additionally, we also incorporate a Temporal Dilated Convolution (TC) block \cite{oord2016wavenet} to extract local information that captures temporal dependency and patterns around the observed measurements. Unlike conventional convolution with limited receptive field, which emphasizes detailed, high-frequency information \cite{wang2020high} and may introduce noise into the model due to the sparsity of IRTS. In contrast, the context of IRTS reveals sparse observations along the timeline, which manifest as low-frequency signals. To address this, we employ a technique known as dilated convolution, a variant of the conventional convolution operator. Dilated convolution inserts "holes" between kernel elements. This design allows the convolution process to selectively disengage from high-frequency signals, thereby enhancing its ability to capture and emphasize low-frequency components within the data and empower the convolution with wider receptive field than standard convolution \cite{yu2017dilated}. Specifically, a multi-scale incremental dilated convolution with $d$ dilation rate (e.g. 1, 2, 3), window size 10, is utilized in our proposed method. The output from the last dilated convolution will be sequentially input to the next dilated convolution. The output from the last convolution layer is then sent to a gated mechanism for adaptively constructing locality representation $e^c_{i} \in \mathbb{R}^{L \times E}$ as shown in Eq. \ref{eq:TC block}. TC block accepts $Tanh$ function as activation function, mapping the input to range $[-1, 1]$ to represent much more complex non-linearity patterns. The $Sigmoid$ function generates a series of values in $[0, 1]$ as a gate to adaptively select the amount of information needed for the problem of interests. 

\begin{equation}
    \mathbf{e}^c_i = tanh(e^c_{d, i}) \odot sigmoid(e^c_{d, i})  \label{eq:TC block}
\end{equation}

\begin{equation}
    \mathcal{F}_i = [\mathbf{e}^c_i || \mathbf{e}^t_i || \mathbf{e}^s_i]
\end{equation}

\begin{table*}[]
    \centering
    % \resizebox{7cm}{1.2cm}{%
    \caption{Statistics of IRTS dataset}
    \begin{tabular}{cccccccc}
         Datasets & \#Samples & \#Sensor & \#Classes & Imbalanced & Missing Ratio & Train:Val:Test & Dataset Type\\
         \hline
         P12 & 11988 & 36 & 2 & True & 88.4\% & 8:1:1 & NIRTS\\
         P19 & 38803 & 34 & 2 & True & 94.9\% & 8:1:1 & NIRTS\\
         PAM & 11988 & 17 & 8 & False & 60.0\% & 8:1:1 & AIRTS\\
         \hline
    \end{tabular}
    % }
    
    \label{tab:dataset_statistics}
\end{table*}

\begin{table*}
    \centering
    \caption{Classic Classification on General IRTS Dataset}
    \begin{tabular}{c|cc|cc|cccc}
        \hline
         \multirow{2}{*}{\centering \textbf{Methods}} & \multicolumn{2}{c|}{P12} &  \multicolumn{2}{c|}{P19} & \multicolumn{4}{c}{PAM} \\
         &  AUC & AUPR & AUC & AUPR & Accuracy & Precision & Recall & F1 Score\\
         \hline
         Transformer & 83.3 \textpm 0.7 & 47.9 \textpm 3.6 & 80.7 \textpm 3.8 & 42.7 \textpm 7.7 & 83.5 \textpm 1.5 & 84.8 \textpm 1.5 & 86.0 \textpm 1.2 & 85.0 \textpm 1.3  \\
         Trans-Mean & 82.6 \textpm 2.0 & 46.3 \textpm 4.0 & 83.7 \textpm 1.8 & 45.8 \textpm 3.2 & 83.7 \textpm 2.3 & 84.9 \textpm 2.6 & 86.4 \textpm 2.1 & 85.1 \textpm 2.4  \\
         GRU-D & 81.9 \textpm 2.1 & 46.1 \textpm 4.7 &  83.9 \textpm 1.7 & 46.9 \textpm 2.1 & 83.3 \textpm 1.6 & 84.6 \textpm 1.2 & 85.2 \textpm 1.6 & 84.8 \textpm 1.2  \\
         SeFT & 73.9 \textpm 2.5 & 31.1 \textpm 4.1 & 81.2 \textpm 2.3 & 41.9 \textpm 3.1 & 67.1 \textpm 2.2 & 70.0 \textpm 2.4 & 68.2 \textpm 1.5 & 68.5 \textpm 1.8  \\
         mTAND & \underline{84.2 \textpm 0.8} & 48.2 \textpm 3.4 & 84.4 \textpm 1.3 & 50.6 \textpm 2.0 & 74.6 \textpm 4.3 & 74.3 \textpm 4.0 & 79.5 \textpm 2.8 & 76.8 \textpm 3.4  \\
         IP-Net & 82.6 \textpm 1.4 & 47.6 \textpm 3.1 & 84.6 \textpm 1.3 & 38.1 \textpm 3.7 & 74.3 \textpm 3.8 & 75.6 \textpm 2.1 & 77.9 \textpm 2.2 & 76.6 \textpm 2.8 \\
         $DGM^2$-O & 83.8 \textpm 0.6 & \underline{48.4 \textpm 2.5} & 85.9 \textpm 3.9 & 41.8 \textpm 10.3 & 82.4 \textpm 2.3 & 85.2 \textpm 1.2 & 83.9 \textpm 2.3 & 84.3 \textpm 1.8  \\
         MTGNN & 74.4 \textpm 6.7 & 35.5 \textpm 6.0 & 81.9 \textpm 6.2 & 39.9 \textpm 8.9 & 83.4 \textpm 1.9 & 85.2 \textpm 1.7 & 86.1 \textpm 1.9 & 85.9 \textpm 2.4  \\
         Raindrop & 82.8 \textpm 1.7 & 44.0 \textpm 3.0 & 87.0 \textpm 2.3 & 51.8 \textpm 5.5 & 88.5 \textpm 1.5 & 89.9 \textpm 1.5 & 89.9 \textpm 0.6 & 89.9 \textpm 1.0  \\
         ContiFormer & 81.2 \textpm 0.8 & 43.9 \textpm 3.0 & 79.2 \textpm 2.3 & 35.8 \textpm 2.3 & 89.0 \textpm 1.0 & 90.0 \textpm 0.8 & 91.0 \textpm 0.9 & 90.2 \textpm 0.8  \\
         WarpFormer & 83.5 \textpm 1.9 & 45.1 \textpm 3.5 & \underline{87.7 \textpm 3.2} & \underline{53.4 \textpm 6.4} & \underline{93.5 \textpm 1.0} & \underline{94.5 \textpm 0.9} & \underline{94.0 \textpm 0.9} & \underline{94.2 \textpm 0.8}  \\
         MTSFormer & \textbf{84.9 \textpm 1.4}  & \textbf{51.1 \textpm 3.7}  & \textbf{88.8 \textpm 1.5} & \textbf{57.7 \textpm 4.4} & \textbf{96.8 \textpm 0.9} & \textbf{97.3 \textpm 0.8} & \textbf{96.9 \textpm 0.6} & \textbf{97.1 \textpm 0.7} \\
        \hline
    \end{tabular}
    
    \label{tab:classic classification}
\end{table*}

\subsection{Multi-Views Self-Attention and Fusion}
After obtaining the individual distinct embedding $e^c_{i}, e^t_i, e^s_i$ from Sensor as Channel, Time as Token, and Sensor as Token, we wish to find the mutual information between the views and preserve their own characteristics by exchanging the information. One promising approach is cross-attention mechanism to exchange information and fuse the information among the views\cite{zhang2023improving}. However, cross-attention only fuses the information between two views individually. Instead, we use multi-view self-attention to learn mutual information of three views at once in space $\mathbb{R}^E$ as shown in Figure. \ref{fig:overall-architecture}. This multi-view self-attention has a more comprehensive understanding of IRTS than cross-attention due to the interaction from all views instead of two individual views. To approach this, we concatenate, denoted as $||$, the learned embeddings from three views, $\mathbf{e}^c_i, \ \mathbf{e}^t_i \in \mathbb{R}^{L \times E}$ and $\mathbf{e}^s_i \in \mathbb{R}^{N_s \times E}$ to form a fused representation $\mathcal{F}_i \in \mathbb{R}^{(2L+N_s) \times E}$ of the sample $\mathcal{X}_i$ and implement muti-view self-attention on $\mathcal{F}_i$ as shown in Eq. \ref{eq:Attention}. Unlike the conventional attention mechanism, focusing on the time dimension, multi-view self-attention integrates indicative views and re-distributes the features of each view by their underlying cross and self-interaction.

$\hat{\mathcal{F}}_i$ is a complicated representation of sample $\mathcal{X}_i$ from three different views with inconsistent measurements. The features of the three views are "re-weighted" after self-attention. Some features in the hidden state can be adjusted to a lower value and less likely to activate ReLU function, resulting in gradient vanishing. In addition, we wish self-attention to find the dynamics between views discarding their unique representation. To address this, a residual connection is employed after self-attention that helps the model to focus on dynamics between views without concern about how much information is to be preserved from the original state. To stabilize the data distribution in the hidden space, a LayerNorm on $\hat{\mathcal{F}}_i$ to mitigate gradient vanishing and discrepancies between views as shown in Eq. \ref{eq:LayerNorm}. Unlike the typical layer normalization on time tokens, layer normalization across different views preserved their mutual information in means and variance, implicitly fusing the global and local tokens behind the scenes, resulting in sophisticated and informative representation. $\hat{\mathcal{F}}_i$ is then passed to a feedforward block to learn the intrinsic property and internal relationship of any input IRTS, such as outstanding signals. After obtaining the information from other views, $\hat{\mathcal{F}}_i$ contains sophisticated and comprehensive information from other views. Then, $\hat{\mathcal{F}}_i$ will be split into three paths to the next layer to find the interaction between individual and fused information in the path view space and make the module stackable for more layers to fit large-scale datasets.

\begin{equation}
    \small 
    \hat{\mathcal{F}}_i = SelfATT(\mathcal{F}_i) = softmax \left( \frac{(W_Q \mathcal{F}_i)  (W_K \mathcal{F}_i)^T }{\sqrt{d_k}} \right) W_V \mathcal{F}_i \label{eq:Attention}
\end{equation}

\begin{equation}
\small 
    LayerNorm(\hat{\mathcal{F}}_i) = \left\{ \frac{\hat{\mathcal{F}}_i - Mean(\hat{\mathcal{F}}_i)}{\sqrt{Var(\hat{\mathcal{F}}_i})} \biggr\rvert i = 1, \dots, N \right\} \label{eq:LayerNorm}
\end{equation}

\subsection{Adaptive Irregularity Learning}
NIRTS often exhibit discernible periodic patterns that mirror the underlying behaviour of the sampled $\mathcal{X}_i$. These patterns can provide valuable insights into the nature of the system being monitored. On the other hand, the missingness patterns in AIRTS data tend to follow a standard Gaussian noise distribution, which can obscure meaningful patterns, complicate predictive modelling efforts and contribute no positive impacts to predictions. Although the original observation $\mathcal{X}_i$ has displayed irregular patterns, it can be biased to make the decisions by focusing on the value itself instead of irregularity. While NIRTS datasets (e.g., ICU monitoring systems) may show discernible periodic patterns in vital indicator values, even among dying patients of different ages and genders, this is not necessarily true for AIRTS datasets. We also empirically approved that the irregularity module improves model performance in the NIRTS dataset (e.g. P12 and P19, refers to Section \ref{sec:dataset}) and degrades performance in AIRTS dataset (e.g. PAM, refers to Section \ref{sec:dataset}) by simply concatenating the inverse mask to the original data and removing it totally from the model.

To address the challenge of adapting to both NIRTS and AIRTS datasets, MTSFormer proposes an irregularity gate to adaptively select information from the irregularity module as shown in Figure \ref{fig:overall-architecture}. This gate aims to empower the model adapting to general types of datasets without any prior knowledge. Specifically, we represent the irregular patterns by a binary mask denoted as irregularity tokens $m^t_{i} \in \mathbb{R}^{L \times N_s}$ for time token and $m^s_{i} \in \mathbb{R}^{L \times N_s}$ for sensor token, generated from the three path original data, where 0 denotes the absence of data measurement and 1 denote the presence. These masks are generated by filling missing values with 0 and replacing present values with 1. Unlike the previous work, simple concatenation of the mask may degrade performance in AIRTS as the mask displays random and misleading information while improving performance for NIRTS. We hypothesise the improved performance comes from the debiased effects of irregular masks that removed the measurement biases in the dataset. Intuitively, measurement biases can be produced by inconsistent normal values of the physical measurements of patients in different age and gender groups \cite{national1990national}.

The encoder of each irregularity mask shares weights with the first layer encoder of each view: $L^1_{tc}$, $L^1_{tt}$, and $L^1_{ts}$. This weight-sharing mitigates value biases caused by inconsistent measurements in the original observations, resulting in a more adaptive and generalized model for both NIRTS and AIRTS datasets. The corresponding mask will be encoded into irregular embeddings $e^c_{m,i}$, $e^t_{m,i}$ and $e^s_{m,i}$ for local irregularity, global irregularity in time dimension, and global irregularity in sensor dimension respectively. To enable adaptive learning and the selection of task-relevant information, $e^c_{m,i}$, $e^t_{m,i}$ and $e^s_{m,i}$ are activated by a Tanh function and then pass to a gate with sigmoid function. The gate provides dynamic control over the flow of information, allowing the model to adaptively focus on the most relevant irregularity for the task in NIRTS and suppress the noisy missingness in AIRTS. Finally, the selected irregularity is incorporated into the final stage embedding from each view with addition operation.

\section{Experiments}
In this section, we conducted extensive empirical study and extensively the impacts of each component of our proposed method. Additionally, we tested different variants of MTSFormer to further explore impacts of irregularity mask on NIRTS and AIRTS dataset to verify our claims in the introduction. In summary, we compared the SOTA baselines specialized on irregular time series data and time series forecasting. MTSFormer significantly improved the performance over all baseline methods in classic classification task and our challenging experiment setting.

\subsection{Dataset and Metrics} \label{sec:dataset}
We implemented experiments on three benchmark datasets: P12, P19 and PAM preprocessed by Raindrop \cite{zhang2021graph}, containing 11988, 38803 and 11988 samples respectively. P12 \cite{citi2012physionet} comprises the data with 36 sensors, 88.4\% missing ratio and binary labels to predict mortality of ICU patients. P19 \cite{reyna2020early} comprises the data with 34 sensors, 94.9\% missing ratio and binary labels to predict sepsis. Notably, P12 and P19 are the natural irregular datasets from real-world medical electronic records, which can be recognized as NIRTS dataset since each sensor is sampled by different frequencies. PAM \cite{reiss2012introducing}, on the other hand, is randomly sampled to be irregular with 60\% missing ratio and consists of 11988 samples, used to simulate the NIRTS generated by randomly systematic fault. The task is to classify 8 distinct human activities from 17 irregularly observed sensors. To be consistent, we employed the same evaluation metrics from Raindrop \cite{zhang2021graph}, Area Under ROC Curve (AUC) and Area Under Precision-Recall Curve (AUPR) for P12 and P19, and Accuracy, Precision, Recall and F1 score for PAM.

\subsection{Baselines}
We compared our method with 11 popular baseline models implemented from previous works: \textbf{Transformer} \cite{vaswani2017attention} and \textbf{Transformer-Mean} that utilize, the benchmark model for many sequential and classification tasks. The Transformer-Mean imputed the missing value with the mean of the sensors along the time dimension and predicts the imputed sample using vanilla Transformer. \textbf{GRU-D} \cite{che2018recurrent} incorporated the Decay mechanism into GRU structure. \textbf{IP-Net} \cite{shukla2019interpolation} interpolates the time series and utilizes GRU as a learning module. \textbf{mTAND}\cite{shukla2021multi} utilize the time attention mechanism to fuse the temporal relationship. As mTAND has shown outstanding performance over Neual ODE\cite{chen2018neural}, ODE-RNN\cite{rubanova2019latent}, we will not include them into the performance comparison. Instead, we approach to compare a strong ODE work, \textbf{ContiFormer} \cite{chen2024contiformer}, that leverages neural ODE with attention mechanism in Transformer architecture. In addition, we also compared our work with the representative graph models, \textbf{MTGNN}\cite{wu2020connecting} and \textbf{Raindrop}\cite{zhang2021graph} designed for IRTS data forecasting and classification. Moreover, we compare our work with other diverse methods, \textbf{SeFT} \cite{horn2020set} that utilize set function, $\mathbf{DGM^2-O}$ \cite{wu2021dynamic}, a Gaussian mixture generated method for imputation, \textbf{WarpFormer} \cite{zhang2023warpformer}, a Transformer variant leveraging Dynamic Time Warpping for measurement unifying and multi-scaled attention mechanism for IRTS. We noticed that ViTST \cite{li2024time} is also designed for IRTS classification, which converts time series data into images and transfers the knowledge from large pre-tained models to IRTS. However, we noticed the performance of ViTST largely depends on the dataset scale that was used to pretrain the Vision Transformer. The ViTST trained from scratch exhibits an unacceptable performance and is even lower than a one-layer vanilla Transformer. Therefore, we do not include ViTST in the comparison to ensure a fair comparison.

\begin{figure*}
     \centering
     \begin{subfigure}[t]{0.185\linewidth}
         \centering
         \includegraphics[width=\textwidth]{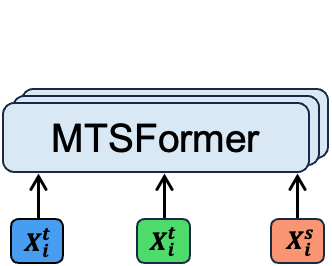}
         \caption{MTSFormer V1}
         \label{fig:MTSFormer V1}
     \end{subfigure}
     % \hfill
     \begin{subfigure}[t]{0.185\linewidth}
         \centering
         \includegraphics[width=\textwidth]{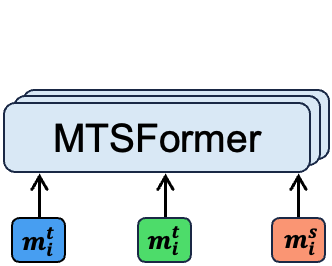}
         \caption{MTSFormer V2}
         \label{fig:MTSFormer V2}
     \end{subfigure}
     % \hfill
     \begin{subfigure}[t]{0.202\linewidth}
         \centering
         \includegraphics[width=\textwidth]{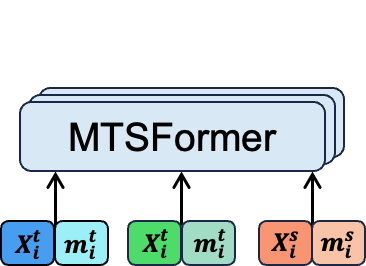}
         \caption{MTSFormer V3}
         \label{fig:MTSFormer V3}
     \end{subfigure}
     % \hfill
     \begin{subfigure}[t]{0.24\linewidth}
         \centering
         \includegraphics[width=\textwidth]{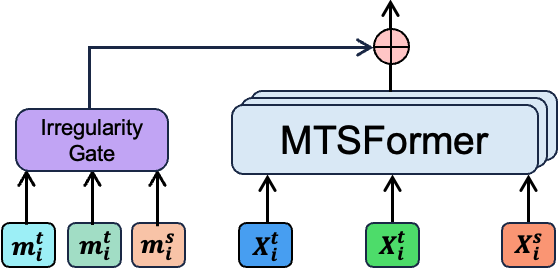}
         \caption{MTSFormer V4}
         \label{fig:MTSFormer V4}
     \end{subfigure}
     
     \caption{MTSFormer Variants}
     \label{fig:MTSFormer Variants}
\end{figure*}

\begin{figure*}
     \centering
     \begin{subfigure}[t]{0.25\linewidth}
         \centering
         \includegraphics[width=\textwidth]{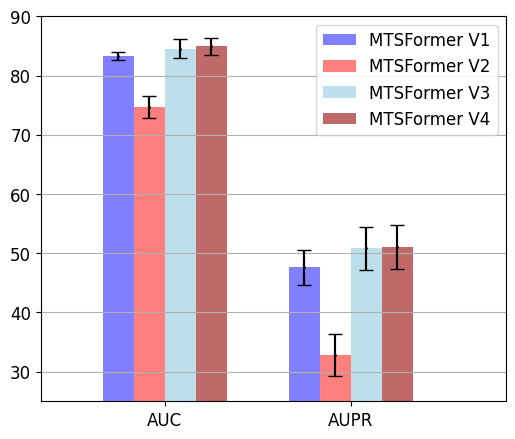}
         \caption{P12}
         \label{fig:P12_Ablation}
     \end{subfigure}
     % \hfill
     \begin{subfigure}[t]{0.25\linewidth}
         \centering
         \includegraphics[width=\textwidth]{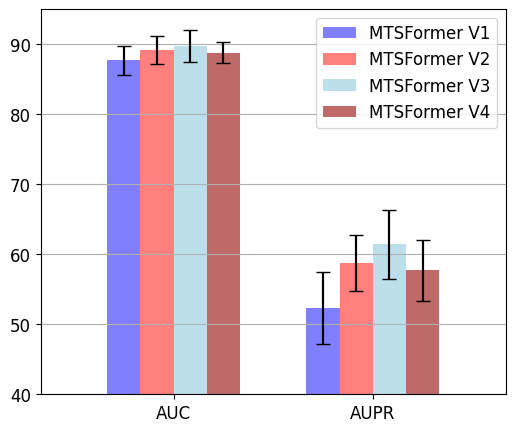}
         \caption{P19}
         \label{fig:P19_Ablation}
     \end{subfigure}
     % \hfill
     \begin{subfigure}[t]{0.415\linewidth}
         \centering
         \includegraphics[width=\textwidth]{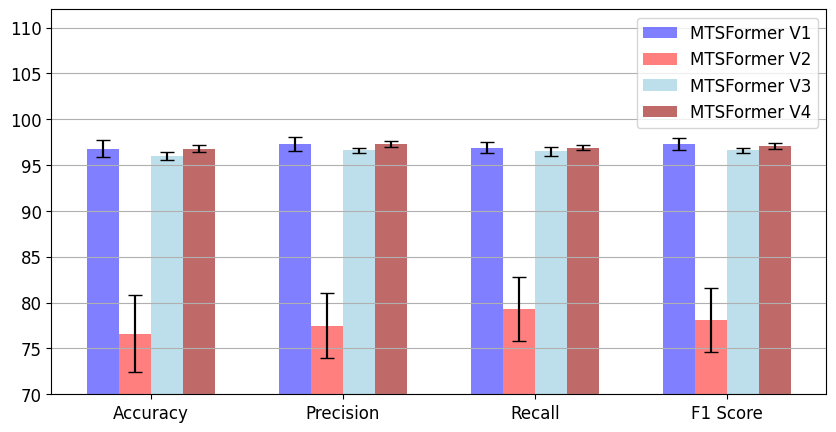}
         \caption{PAM}
         \label{fig:PAM_Ablation}
     \end{subfigure}
     
     \caption{Variants Experiments on general IRTS dataset}
     \label{fig:Variants Ablation}
\end{figure*}

\subsection{Main Results}
We implement experiments with full data usage in P12, P19, PAM and the parameters of baseline models: Transformer, GRU-D, SeFT, mTAND, IP-Net, $DGM^2-O$, MTGNN, Raindrop are kept from the Raindrop repository \cite{zhang2021graph}. ContiFormer and WarpFormer are kept from their recommended parameters. To ensure a fair comparison, all the results are obtained from 5-fold cross-validation. 

The results are reported by mean \textpm  std as shown in Table. \ref{tab:classic classification}. Our proposed method demonstrates the highest performance against all baseline methods from graph-based, Transformer-based, and specialized methods. For those NIRTS dataset, MTSFormer improved over the strongest baseline, WarpFormer, by 1.4\% AUC and 6\% AURP on P12, 1.1\% AUC and 4.3\% AUPR on P19. For PAM dataset, the improvement is more significant, with increases of 3.3\% on Accuracy, 3.2\% on Precision, 2.9\% on Recall and 3.1\% on F1 score. It is also worthy to note that our method exhibits more stable performance than WarpFormer and higher performance upper bond. The outstanding performance indicated that MTSFormer has the highest robustness to both NIRTS and AIRTS scenarios. Although ContiFormer is theoretically strong in modelling IRTS, it fails to handle natural irregular data since the performance in P19 and P12 is relatively low. Especially in P19, the result of Contiformer is lower than vanilla Transfromer. Those graph-based models, such as MTGNN and Raindrop, also fails on P12 dataset, indicating the graph dependency relationship is not the key factor for IRTS classification task. Interestingly, we also reproduced the results of ViTST \cite{li2024time} that leveraged the reasoning ability of large pre-trained vision model, 85.3\% AUC and 48.1\% on P12, 89.5\% AUC and 52.2\% on AUPR on P19, 95.8\% Accuracy, 96.2\% Precision, 96.1\% Recall, 96.5\% F1 score on PAM. We obtained the above results by a randomly generated seed since the reported results of ViTST are produced by setting a specific seed. Our proposed method also wins ViTST the AUPR scores on both P12 and P19 and slightly improve all metrics of PAM dataset. However, the promising performance from ViTST came from the pre-trained ViTST. This approach is failed on modelling IRTS from scratch. For this reason, we exclude this method for formal comparison.

Notably, mTAND, $DGM^2-O$ and Raindrop display distinguished performance in P12 and P19, the NIRTS data, ranking just below our proposed method. However, they are unable to withstand other baselines with powerless performance in PAM, the AIRTS dataset. Similarly, ContiFormer, WarpFormer displayed competitive results in PAM with the third and second ranking below MTSFormer and incapable learning ability in P12. Our proposed method, on the other hand, outperforms all other baselines with superior results and significant improvements. This provides empirical and strong evidence to support our hypothesis that the enhancements in the model are attributable to the well-crafted design of the irregularity-gated mechanism.

\begin{figure*}[h]
     \centering
     \begin{subfigure}[t]{0.47\linewidth}
         \centering
         \includegraphics[width=\textwidth]{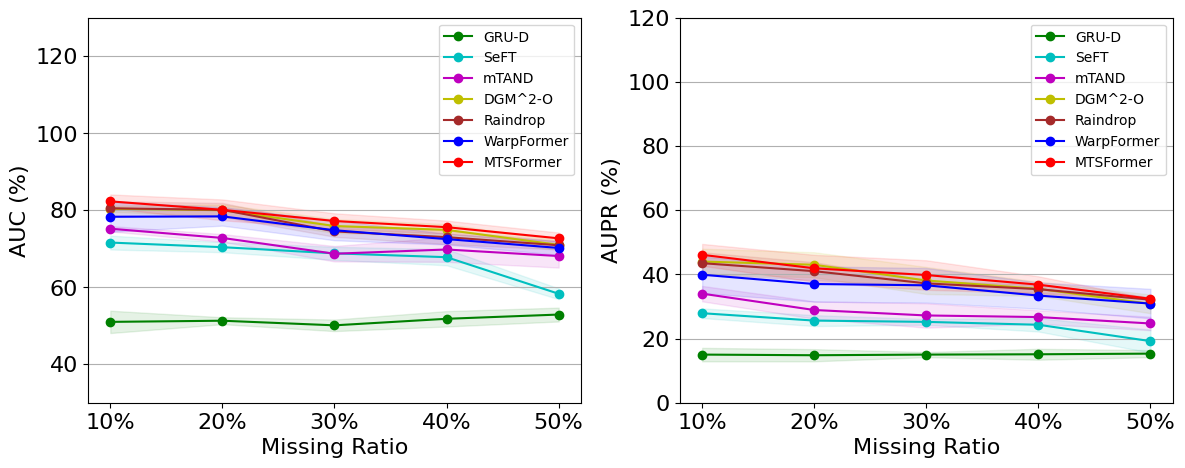}
         \caption{P12}
         \label{fig:P12_Missing_Ratio}
     \end{subfigure}
     % \hfill
     \begin{subfigure}[t]{0.47\linewidth}
         \centering
         \includegraphics[width=\textwidth]{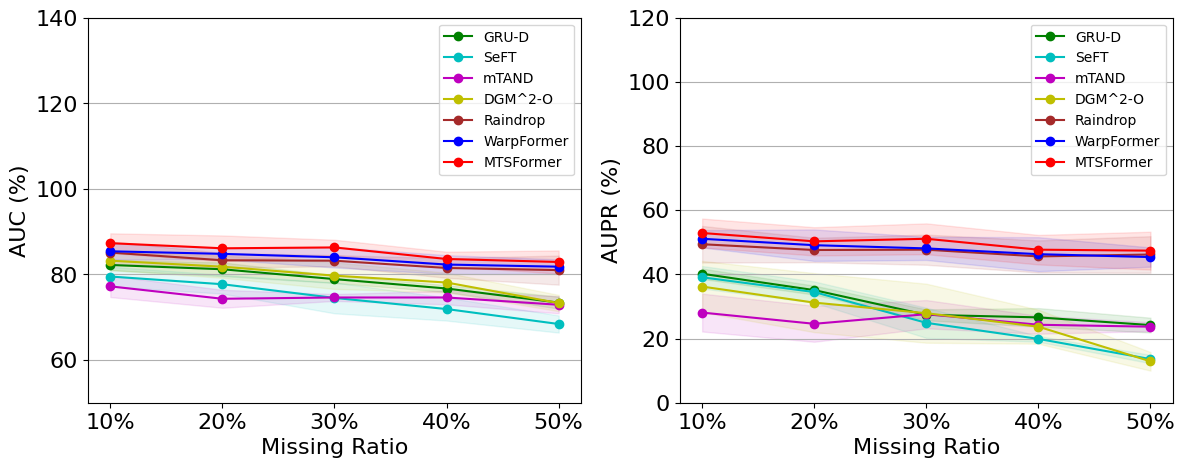}
         \caption{P19}
         \label{fig:P19_Missing_Ratio}
     \end{subfigure}
     % \hspace{0.1}
     \begin{subfigure}[t]{0.951\linewidth}
         \centering
         \includegraphics[width=\textwidth]{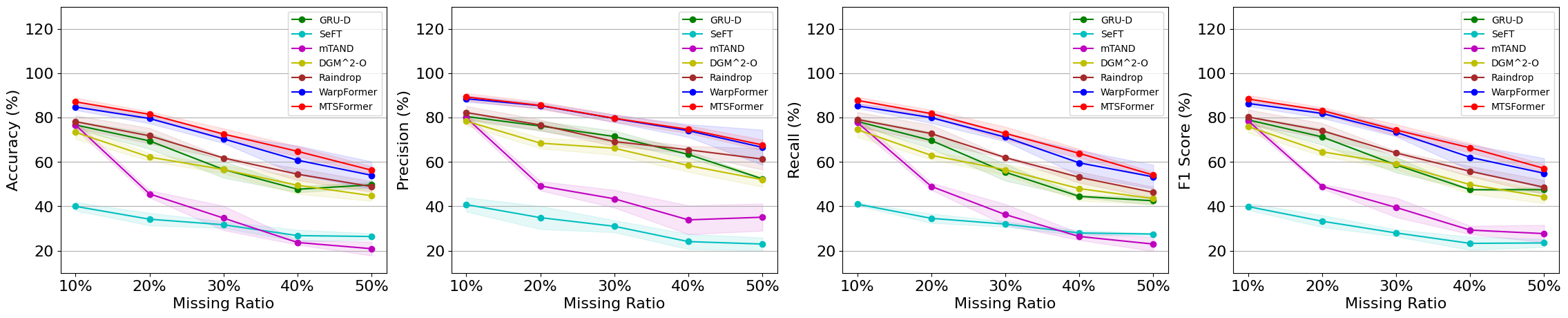}
         \caption{PAM}
         \label{fig:PAM_Missing_Ratio}
     \end{subfigure}
     
     \caption{Leave-Random-Sensor-Out Experiments}
     \label{fig:Leave-random-sensor}
\end{figure*}

\begin{table*}[ht]
    \centering
    \caption{Ablation Study on Each Component}
    \begin{tabular}{c|cc|cc|cccc}
        \hline
        \multirow{2}{*}{\centering \textbf{MTSFormer}} & \multicolumn{2}{c|}{P12} & \multicolumn{2}{c|}{P19} & \multicolumn{4}{c}{PAM} \\
        & AUC & AUPR &  AUC & AUPR & Accuracy & Precision & Recall & F1 score \\
        \hline
        TC & 84.3 \textpm 1.9 & 48.6 \textpm 4.8 & 88.5 \textpm 1.5 & 56.6 \textpm 6.1 & 95.6 \textpm 0.6 & 96.1 \textpm 0.5 & 95.6 \textpm 0.6 & 96.0 \textpm 0.6  \\
        Time & 82.3 \textpm 1.7 & 45.0 \textpm 3.8 & 88.5 \textpm 1.9 & 55.8 \textpm 4.0 & 91.5 \textpm 1.7 & 93.5 \textpm 1.6 & 92.6 \textpm 1.5 & 92.8 \textpm 1.3  \\
        Sensor & 71.0 \textpm 6.3 & 29.5 \textpm 6.3 & 80.7 \textpm 1.4 & 44.3 \textpm 4.5 & 81.8 \textpm 1.2 & 83.8 \textpm 1.3  & 83.8 \textpm 1.6 & 83.3 \textpm 1.2  \\
        IR Mask & 74.1 \textpm 2.0 & 33.3 \textpm 3.5 & \textbf{89.0 \textpm 2.0} & \textbf{58.8 \textpm 4.0} & 76.6 \textpm 4.2 & 77.5 \textpm 3.5 & 79.3 \textpm 3.5 & 78.1 \textpm 3.5  \\
        TC-Time & \underline{84.5 \textpm 1.8} & \underline{50.8 \textpm 4.0} & 88.7 \textpm 1.5 & 56.0 \textpm 4.2 & 94.6 \textpm 0.9 & 95.6 \textpm 0.9 & 95.3 \textpm 0.6 & 95.4 \textpm 0.8  \\
        TC-Sensor & 83.7 \textpm 1.6 & 49.1 \textpm 2.8 & 86.5 \textpm 1.5 & 51.5 \textpm 4.7 & \underline{96.6 \textpm 0.5} & \underline{97.3 \textpm 0.6} & \underline{96.8 \textpm 0.3} & \underline{97.0 \textpm 0.3}  \\
        Time-Sensor & 80.5 \textpm 1.8 & 41.3 \textpm 1.9 & 88.4 \textpm 1.8 & 53.8 \textpm 4.7 & 93.0 \textpm 1.4 &  94.5 \textpm 1.5 & 94.1 \textpm 0.6 & 94.1 \textpm 1.0  \\
        Full & \textbf{84.9 \textpm 1.4} & \textbf{51.1 \textpm 3.7} &  \underline{88.8 \textpm 1.5} & \underline{57.7 \textpm 4.4} & \textbf{96.8 \textpm 0.9} & \textbf{97.3 \textpm 0.8} & \textbf{96.9 \textpm 0.6} & \textbf{97.1 \textpm 0.7}  \\
        \hline
    \end{tabular}
    \label{tab:ablation study}
\end{table*}

\subsection{Leave Random Sensor Out}
NIRST and AIRTS dataset are commonly observed in real-world scenarios. However, some sensors may be deactivated across the timeline due to systematic failure. To simulate such scenarios, we implement Leave-Random-Sensor-Out setting experiments same as Raindrop \cite{zhang2021graph} on both NIRTS and AIRTS. The results are presented in Figure. \ref{fig:Leave-random-sensor}. Specifically, we randomly choose a percentage subset of sensors on test and validation data and replace the chosen sensor readings with all zeros while the model is trained on complete data without any data dropped. We conducted the experiments with a variety range of Leave-Random-Sensor-Out ratio on NIRTS dataset (P12, P19) to simulate accidental sensor failure, making them more challenging than PAM dataset with only 60\% missing ratio. We found that MTSFormer achieves better results on NIRTS on both P12 and P19 than the most competitor, WarpFormer and Raindrop. Notably, our method always displays outstanding AUPR scores over all baselines, indicating high robustness on handling imbalanced data. It is worthy to notice that the temporal method GRU-D barely learn anything from P12 dataset, suggesting that modelling IRTS is challenging for temporal base models. Apart from NIRTS dataset, we implement the same setting on PAM dataset, which are significantly improved to the most competitive baseline, WarpFormer, on Accuracy, Recall and F1 scores. These significant results indicated that our method not only robust to normal NIRTS and NIRTS dataset, but also displays strong resistance to missing sensors on both NIRTS and NIRTS datasets.

\subsection{Effects of Irregularity Pattern}
To further explore the impacts of irregularity, we implement different variants of MTSFormer on P12, P19, and PAM. All parameters are kept the same as in the classic classification experiments in Table .\ref{tab:classic classification}. The architecture of variants are shown in Figure. \ref{fig:MTSFormer Variants}. MTSFormer V1 simply inputs the original value $x^t_i$ and $x^s_i$ into three different paths. MTSFormer V2 replaces the original time series input $x^t_i$ and $x^s_i$ with their corresponding irregularity mask $m^t_i$ and $m^s_i$ and mute the irregular gate. MTSFormer V3 concatenates the irregularity mask with their corresponding value of time series. The concatenation of the irregularity mask in MTSFormer V3 directly accepts the irregularity patterns without diminishing of information. MTSFormer V4 is our proposed method that employs the gate mechanism to adaptively select irregularity information for any generalized dataset. 

We reported the results in Figure. \ref{fig:Variants Ablation}. The highest performance variant is MTSFormer V4 on P12, MTSFormer V3 on P19 and MTSFormer V1 on PAM. By comparison with MTSFormer V1, MTSFormer V3 significantly improved AUC and AUPR performance on NIRTS P19 and P12 dataset with concatenation of mask, empirically demonstrating that irregularity patterns can positively contribute to the modelling of natural irregular time series. Although this concatenation demonstrates impressive effectiveness on P19, full irregular information from irregular masks may generalize poorly on other dataset such as P12 and PAM by introducing noise and task-irrelevant information. Instead, our proposed method MTSFormer V4 generalizes well on three datasets and outperforms all the variants on P12 by the novel design of irregular gate mechanism. On the other hand, P19 demonstrated high correlation with irregular patterns proved by the results of MTSFormer V2. It is impressive that the performance of MTSFormer V2 is even higher than MTSFormer V1 and MTSFormer V4. MTSFormer V2 performs as poor as expected on PAM dataset since the irregular patterns of PAM is randomly sampled. This is strong experimental evidence to support our hypothesis that irregular patterns are highly correlated to model time series in NIRTS dataset but not in AIRTS dataset. Moreover, MTSFormer V1 is the highest-performance model on PAM by compared to others. The second largest is MTSFormer V4 followed by MTSFormer V3. We can conclude that irregular masks do not positively contribute to performance in total but may degrade the performance in the AIRTS dataset. Therefore, the gated mechanics dynamically filter out task-irrelevant information and preserve useful information for decision-making, achieving a performance close to MTSFormer V1.

Although our proposed method, MTSFormer V4 is not always the best model among the variants, the dynamic gated mechanism for irregular mask empowers MTSFormer V4 to produce unbiased prediction and the best generalization performance to both NIRTS and AIRTS datasets. This ability makes our proposed method highly generalized to wider range scenarios of irregular time series data.

\subsection{Ablation Study}
To further investigate the importance of each component in our framework, we conducted an additional ablation study, the results of which are displayed in Table \ref{tab:ablation study}. We tested our components with individual active modes. For example, "TC" denotes that only the temporal convolution path and the temporal mask with gate are activated, while the other paths are muted from the model. The "IR mask" represents the same model configuration and setting as MTSFormer V2 mentioned above, and we displayed its performance here for comparison.

Among the individual views, TC emerged as the strongest performer, achieving the best performance across all three datasets, closely followed by the Time view. In contrast, the Sensor view displayed the weakest results when operating individually, likely due to its lack of temporal relation to neighbouring time steps and lack of irregular patterns over time dimension. On the other hand, Sensor view works well with PAM dataset, in which the irregular patterns have low correlation, also proven by the results of TC-Sensor and Time-Sensor variants that improve the performance of individual TC and Time settings on PAM dataset. This suggests that the sensor view path may not align well with NIRTS due to its sparsity strong correlation to temporal information. Impressively, we tested only the binary IR Mask on three different datasets and obtained the best results on the P19 dataset. By comparison with MTSFormer V1 on P19 shown in Figure. \ref{fig:Variants Ablation}, the distinct difference in AUPR strongly supports our claims that the irregularity itself contains rich information and the original value may be biased due to inconsistent and wide range of measurement. empirically demonstrated the high correlation to irregular patterns based on the performance of IR Mask, which indicates the insights that the original value can generate biased prediction due to inconsistent measurement and wider range of readings.

\section{Conclusion}
In this paper, we introduce a novel multi-modal framework for general irregular time series data. The proposed method incorporates multiple views from time series data and adaptively fuses irregularity patterns by a gate mechanism. This method generalized well on both natural irregular and accidental irregular time series data. Through extensive experiments, our approach demonstrated the strongest performance over both NIRTS and AIRTS dataset. We also demonstrate that our approach surpasses SOTA methods for irregular time series and maintains strong resistance under various missing ratio settings. Surprisingly, we found the irregularity is sufficiently informative in modelling NIRTS along with only mask. Furthermore, we studied the impacts of irregular patterns and discussed in-depth about contribution and necessity of each component in our design.

\section{Acknowledgement}
This work was supported by Australian Research Council Linkage (Grant No. LP230200821), Australian Research Council Discovery Projects (Grant No. DP240103070), Australian Research Council ARC Early Career Industry Fellowship (Grant No. IE230100119), Australian Research Council ARC Early Career Industry Fellowship (Grant No. IE240100275), University of Adelaide, Sustainability FAME Strategy Internal Grant 2023.

\bibliography{aaai25}
\end{document}